\documentclass[10pt, a4paper]{article}

\usepackage{enumitem}
\usepackage{rotating,caption}
\usepackage{todonotes}
\usepackage{multirow}
\usepackage{booktabs}
\usepackage{amsmath}
\usepackage{graphicx}
\usepackage{tikz}
\usetikzlibrary{shadows}
\usetikzlibrary{arrows}
\usepackage{tabularx}
\usepackage{ragged2e}
\usepackage{colortbl}
\usepackage{wrapfig}
\newcommand\posscite[1]{\citeauthor{#1}'s (\citeyear{#1})}

\usepackage[]{lrec-coling2024} 

\title{AQuA - Combining Experts’ and Non-Experts’ Views To Assess Deliberation Quality in Online Discussions Using LLMs}

\name{\begin{tabular}{c}
Maike Behrendt\textsuperscript{1}, Stefan Sylvius Wagner\textsuperscript{1}, Marc Ziegele\textsuperscript{1}, \\ Lena Wilms\textsuperscript{1}, Anke Stoll\textsuperscript{2}, Dominique Heinbach\textsuperscript{3}, Stefan Harmeling\textsuperscript{4}\end{tabular}}
\address{\begin{tabular}{c}\textsuperscript{1}Heinrich Heine University Düsseldorf, Germany 
\textsuperscript{2}Technical University Ilmenau, Germany\\\textsuperscript{3} Johannes Gutenberg University Mainz, Germany \textsuperscript{4}Technical University Dortmund, Germany\end{tabular}\\
         \textsuperscript{1}\{maike.behrendt, stefan.wagner, lena.wilms, marc.ziegele\}@uni-duesseldorf.de\\
         \textsuperscript{2}anke.stoll@tu-ilmenau.de \textsuperscript{3}dominique.heinbach@uni-mainz.de
         \textsuperscript{4}stefan.harmeling@tu-dortmund.de}

\abstract{
    Measuring the quality of contributions in political online discussions is crucial in deliberation research and computer science. Research has identified various indicators to assess online discussion quality, and with deep learning advancements, automating these measures has become feasible. While some studies focus on analyzing specific quality indicators, a comprehensive quality score incorporating various deliberative aspects is often preferred. In this work, we introduce AQuA, an additive score that calculates a unified deliberative quality score from multiple indices for each discussion post. Unlike other singular scores, AQuA preserves information on the deliberative aspects present in comments, enhancing model transparency. We develop adapter models for 20 deliberative indices, and calculate correlation coefficients between experts' annotations and the perceived deliberativeness by non-experts to weigh the individual indices into a single deliberative score. We demonstrate that the AQuA score can be computed easily from pre-trained adapters and aligns well with annotations on other datasets that have not be seen during training.  The analysis of experts' vs. non-experts' annotations confirms theoretical findings in the social science literature.
\\ \newline  }

\begin{document}

\maketitleabstract

\section{Introduction}
In the evolving landscape of democratic discourse, the concept of deliberation stands as a cornerstone, embodying the exchange of ideas, critical discussion, and consensus-building among citizens \citep{dryzek2002deliberative}. Central to the efficacy of these deliberations is their quality, a multifaceted construct traditionally gauged by dimensions such as rationality, civility, reciprocity, and constructiveness \citep{friess2015systematic}. 
More recent research has explored various indicators of deliberative quality in online discussions \citep{steenbergen2003measuring,friess2015systematic,scudder2022measuring}. However, most of these approaches require manual annotation of discussion data from trained coders and serve to analyze the discussion in retrospect. 
As the digital age drives an increasing volume of public conversations onto online platforms, the demand to assess their quality through the previously mentioned dimensions in an automated, scalable manner is growing \citep{diakopoulos2015picking,10.1093/oxfordhb/9780190460518.013.23}.

Previous efforts have demonstrated the potential of using natural language processing (NLP) and machine learning algorithms to automatically identify features of deliberation such as argumentative structure, emotional tone, and engagement patterns \citep{10.1162/coli_a_00364,https://doi.org/10.1002/eng2.12189,su13031187}. The interest in automating such assessments, with projects like the one implemented by \citet{falk-lapesa-2023-bridging} in their examination of argument and deliberative quality with adapter models \citep{pmlr-v97-houlsby19a}, is growing.

Motivated by this research, this study introduces AQuA, an index to measure the deliberative quality of individual comments in online discussions with a single score. While there is an ongoing debate on the usefulness of aggregating multiple indices of deliberation \citep{10.1093/oso/9780192848925.003.0006}, we argue that for some tasks a single value, composed of several theoretically based criteria is favorable. 
Our approach combines predictions on various dimensions of deliberation with insights gained from both expert and non-expert evaluations, resulting in a single deliberative quality score. We make use of data that has been annotated from both trained experts and crowd annotators, representing the non-experts' view. We calculate correlation coefficients between the annotated deliberative quality criteria and the perceived deliberativeness of the comments to attribute importance to each individual criterion. 

\paragraph{Our contributions:}
\begin{enumerate}
\item We train 20 adapter models on aspects of deliberation to form the basis for a single deliberation score.
\item To combine the automated predictions in a meaningful way, we calculate the correlation coefficients between experts' and non-experts' assessments of deliberative quality.
\item We define a single normalized score using the correlations as weights, hereby, creating an interpretable and explainable measure for deliberative quality.
\item Finally, we show in experiments that our score can automatically assess the deliberative quality of discussion comments.  
\end{enumerate}

Our method consists of two components: (1) the utilization
of adapters trained on discrete facets of deliberation, 
and (2) the integration of correlations between
annotations from experts and non-experts to establish a 
normalized score for deliberative quality. In developing this index, we extensively test and evaluate its effectiveness across diverse datasets, demonstrating its utility in real-world applications. By doing so, we aim to contribute to the burgeoning field of computational social science, offering scholars, policymakers, and practitioners a tool to monitor and analyze public dialogues. Our trained adapter weights and the code for calculating AQuA scores are available under \url{https://github.com/mabehrendt/AQuA}. 


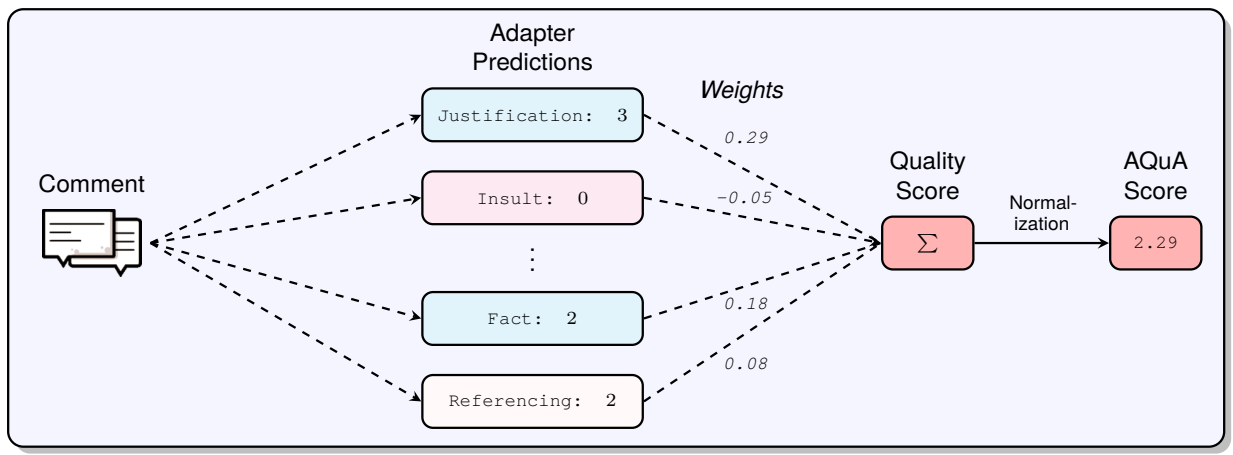
\begin{figure*}
\centering
\begin{tikzpicture}[block/.style={rectangle, draw, line width=0.3mm, minimum width=2.9cm, minimum height=0.7cm, rounded corners, font=\scriptsize\ttfamily}]
    \node[block, minimum width=\textwidth, minimum height=5.8cm, line width=0.3mm, rounded corners=2mm, fill=blue!4, drop shadow] (main) at (-1.3,1.9) {};
    \node[block, fill=cyan!10] (A) at (-2.4,3.4) {Justification: $3$};
    \node[block, fill=magenta!10] (B) at (-2.4,2.3) {Insult: $0$};
    \node[block, fill=cyan!10] (C) at (-2.4,0.7) {Fact: $2$};
    \node[block, fill=pink!10] (D) at (-2.4,-0.4) {Referencing: $2$};
    \node[block, minimum width=1.2cm, fill=red!30] (E) at (2.8,1.7) {$\sum$};
    \node[block, minimum width=1.2cm, fill=red!30] (F) at (5.8,1.7) {2.29};
    \node[font=\scriptsize\itshape\ttfamily] (G) at (-8.2,1.7) {\includegraphics[width=1.3cm]{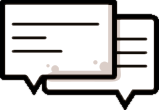}};

    \draw[dashed, line width=0.3mm] (A.east) -- (E.west);
    \draw[-stealth,dashed, line width=0.3mm] (B.east) -- (E.west);
    \draw[dashed, line width=0.3mm] (C.east) -- (E.west);
    \draw[dashed, line width=0.3mm] (D.east) -- (E.west);
    \draw[-stealth,line width=0.3mm] (E.east) -- (F.west);
    \draw[-stealth,dashed, line width=0.3mm] (G.east) -- (A.west);
    \draw[-stealth,dashed, line width=0.3mm] (G.east) -- (B.west);
    \draw[-stealth,dashed, line width=0.3mm] (G.east) -- (C.west);
    \draw[-stealth,dashed, line width=0.3mm] (G.east) -- (D.west);

    \node[font=\footnotesize, text centered, text width=1.5cm] at (-8.2,2.5) {Comment};
    \node[font=\footnotesize, text centered, text width=1.5cm] at (2.8,2.6) {Quality Score};
    \node[font=\footnotesize, text centered, text width=1.5cm] at (5.8,2.6) {AQuA Score};
    \node[font=\scriptsize, text centered, text width=1.5cm] at (4.3,2.1) {Normal-\\ization};
    \node[font=\footnotesize, text centered, text width=2.4cm] at (-2.4,4.3) {Adapter Predictions};
    \node[font=\footnotesize\itshape, text centered, text width=1cm] at (0.3, 3.7) {Weights};

    \node[font=\footnotesize, text centered, text width=1cm] at (-2.4,1.5) {\rotatebox{90}{\dots}};

    \node[font=\scriptsize\itshape\ttfamily, text centered, text width=1cm] at (0.4,3.1) {0.29};
    \node[font=\scriptsize\itshape\ttfamily, text centered, text width=1cm] at (0.4,2.3) {-0.05};
    \node[font=\scriptsize\itshape\ttfamily, text centered, text width=1cm] at (0.4,0.9) {0.18};
    \node[font=\scriptsize\itshape\ttfamily, text centered, text width=1cm] at (0.4,0.1) {0.08};

\end{tikzpicture}
\caption{AQuA calculates a single score for deliberativeness from weighted adapter predictions on 
20 different deliberative aspects. The adapter predictions are weighted by the correlation coefficients between each deliberative aspect and the perception of crowd workers about whether a comment 
is deliberative or not. The normalized score can then be used to compare the deliberative quality of individual comments.}
\label{fig:aqua}
\end{figure*}

\section{Related Work}
Before explaining our approach in detail, we give an overview on the previous work to quantify aspects of deliberation in online discussions and the adapter approach to efficiently train language models for downstream tasks.

\subsection{Deliberative Quality Indices}

Various attempts have been made in the literature to conceptualize deliberation aspects to assess the quality of discourse. Here, we provide a summary of key indicators and metrics proposed in this domain.

The \emph{Deliberative Quality Index} (DQI), introduced by  \citet{steenbergen2003measuring} and further refined by \citet{10.1093/oso/9780192848925.003.0006}, is a prominent and frequently applied metric for evaluating deliberative quality. The DQI comprises five dimensions: \emph{equality of participation}, \emph{level of justification}, \emph{content of justification}, \emph{respect}, and \emph{constructive politics}. These dimensions are assessed for each contribution and averaged for a single speaker.

\posscite{scudder2022measuring} \emph{Listening Quality Index} (LQI) emphasizes deliberative listening as a crucial factor in communication quality, organizing elements of existing measures into a hierarchical scale. This scale progresses from minimal listening to a stage where the speaker feels acknowledged, emphasizing the sequential fulfillment of criteria. The LQI differentiates between speakers and listeners, considering not just the contributions to the dialogue but also the participants' behavior and their feeling of being heard. 

The \emph{Deliberative Reason Index} (DRI) by \citet{NIEMEYER_VERI_DRYZEK_BAECHTIGER_2024} seeks to capture deliberative quality at the group reasoning level rather than evaluating individual contributions. This approach, akin to the LQI, employs surveys conducted before and after discussions to gauge participants' views and preferences on debated topics, calculating agreement scores that are then aggregated to a group score. 

Although referred to as indices, the discussed methodologies do not necessarily provide a single index. They often yield multiple metrics rather than a singular measure, demanding a comprehensive evaluation to determine the overall quality of contributions or debates.  \citet{doi:10.1080/10584609.2020.1830322} suggest aggregating the presence of deliberative qualities — rationality, respect, reciprocity, and civility — and computing their average to establish a quality ratio, treating each criterion with equal importance. We argue, however, that certain aspects may be more important than others to estimate the deliberative quality of a contribution \citep{chen2017online}. 

While the indices presented are valuable for in-depth political debate analysis, their application requires extensive effort from trained coders for annotation and reliability assessments. To streamline the analysis of the deliberative quality of online discussions, several automation proposals have emerged. For instance, \citet{wyss2015decline} employ cognitive complexity to analyze Swiss parliamentary debates, using indicators derived from the Linguistic Inquiry and Word Count (LIWC) dictionary \citep{doi:10.1177/0261927X09351676}. \citet{10.1093/llc/fqv033} automate the measurement and annotation of features like participation and justification, subsequently employing a visual analytics system for data representation. \citet{fournier-tombs_delibanalysis_2020}  introduced DelibAnalysis, a framework for predicting the DQI of online discussion contributions through machine learning, while \citet{su13031187} proposed leveraging network and time-series analyzes to assess deliberation criteria automatically. 

Our proposed method seeks to bridge the gap between NLP techniques and the theoretical aspects of deliberative quality assessment. We introduce the AQuA score to (i) combine the theoretical underpinnings of deliberation with the comment quality in online debates as perceived by non-experts, and thereby (ii) offering a tool to quantify deliberation aspects through advanced deep learning methods.

\subsection{Adapters}
Adapters, as introduced by \citet{NIPS2017_e7b24b11} are an efficient approach to customize pre-trained language models like RoBERTa \citep{liu2019roberta} for specific tasks.  This method involves the integration of additional bottleneck layers into the model for each distinct task, which adds new weights while leaving the original pre-trained weights unaltered.

The concept of adapter layers was first applied to NLP by \citet{pmlr-v97-houlsby19a}, who adapted the Transformer architecture \citep{NIPS2017_3f5ee243} to include these layers. The design of the adapter involves compressing the input's dimensionality to a significantly smaller size, applying a non-linear function, and incorporating a skip-connection to circumvent the bottleneck, with task-specific layer normalization parameters also being adjustable.

The strategic insertion of adapter layers has been a focus of research, with \citet{pmlr-v97-houlsby19a} positioning them subsequent to both the multi-head attention and feed-forward layers within the Transformer architecture. \citet{pfeiffer-etal-2021-adapterfusion} found in an extensive search on architectural parameters, that placing only one adapter after the feed forward layer in the Transformer works best throughout all their experiments. We also apply this architecture for our models.
The introduction of AdapterHub by \citet{pfeiffer-etal-2020-adapterhub} and the adapters library by \citet{poth2023adapters} further facilitated the sharing and reuse of pre-trained adapters within the community.

Subsequent studies, such as those by \citet{mendonca-etal-2022-qualityadapt}, explored the training of individual adapters for dialogue quality estimation, and the use of AdapterFusion \citep{pfeiffer-etal-2021-adapterfusion} to merge features from different adapters. \citet{falk-lapesa-2023-bridging} trained 20 adapters on features for argument and deliberative quality to examine their dependencies. In our work, we follow a similar path to train adapters to evaluate specific aspects of deliberative quality and subsequently combine them using correlation coefficients between experts' and non-experts' annotations, to create a single deliberative quality metric.

\section{AQuA: An Additive Score for Deliberative Quality}

With AQuA we propose a metric for assessing the quality of individual comments in online discussions. Our approach combines predictions on various dimensions of deliberation with insights gained from both experts' and non-experts' evaluations, resulting in a single deliberative quality score. Our methodology consists of two components: (1) the utilization of adapters trained on discrete facets of deliberation, and (2) the integration of correlations between experts' and non-experts' annotations to establish a normalized score for deliberative quality. We therefore harness annotations of the same data, once labeled by trained experts for a variety of deliberative qualities, such as the degree of justification, and once labeled by non-experts on their personal assessment of the deliberativeness of a comment. We calculate correlation coefficients between each individual deliberative criterion (experts' labels) and the binary indicator for deliberativeness (non-experts' labels).

The idea of our approach is to aggregate individual scores calculated by adapters in a meaningful way to obtain a single score for each comment, in which some aspects contribute more to the perceived deliberativeness than others. For this reason we call our approach AQuA, an ``Additive deliberative Quality score with Adapters''.

\subsection{Datasets}
\label{subsec:datasets}

Our analysis is based on three datasets: 
\begin{enumerate}
    \item The KODIE dataset, comprising 13,587 comments that were collected and annotated as part of a  scientific study that explored the impact of news organizations’ interactive moderation on the deliberative quality of users’ political discussions \citep{heinbach2022}.
The comments were posted on the Facebook pages of four German national and regional news outlets with high outreach and diverse audiences. These news outlets delivered data that included all published and deleted/hidden posts and comments on their Facebook pages for a period of 12 weeks per news outlet.
\item
The \#meinfernsehen2021 (German for my television) dataset \citep{2022:gerlach:meinferns} is the result of a large scale citizen participation on the future of public television in Germany. Overall, 1,714 comments from the participation process have been manually coded as part of a quantitative content analysis to examine the discussion quality.
\item The CrowdAnno project \citet{wilms2023} collected a non-expert representation of deliberative quality via crowd annotations for a subset of, i.a., both the KODIE and \#meinfernsehen datasets.
\end{enumerate}

 The annotations from two different perspectives are explained in the following.


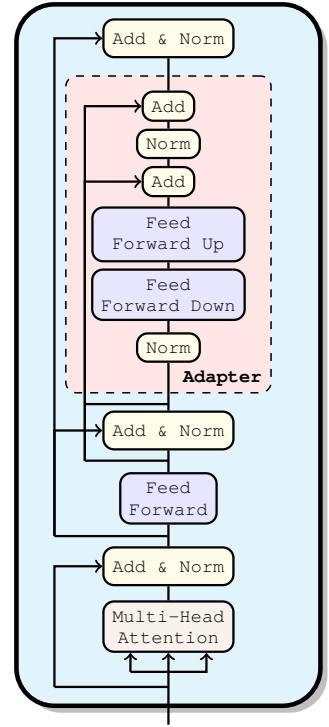
\begin{figure}
    \centering
    \begin{tikzpicture}[block/.style={rectangle, draw, line width=0.3mm, rounded corners, font=\ttfamily\scriptsize}]
        \node[block, minimum width=4.0cm, minimum height=9.3cm, line width=0.6mm, rounded corners=5mm, fill=cyan!10, drop shadow] (main) at (0,-0.6) {};
        \node[block, dashed, minimum width=2.7cm, minimum height=4.2cm, line width=0.2mm, fill=red!10] (main) at (0,1.0) {};

        \node[block, minimum width=1cm, minimum height=0.5cm, line width=0.3mm, align=center, fill=yellow!10] (add3) at (0, 3.6) {Add \& Norm};

        \node[block, draw, minimum width=0.5cm, minimum height=0.2cm, line width=0.3mm, align=center, fill=yellow!10] (addadap2) at (0,2.7) {Add};
        \node[block, draw, minimum width=0.5cm, minimum height=0.2cm, line width=0.3mm, align=center, fill=yellow!10] (normadap2) at (0,2.2) {Norm};
        \node[block, draw, minimum width=0.5cm, minimum height=0.2cm, line width=0.3mm, align=center, fill=yellow!10] (addadap1) at (0,1.7) {Add};

        \node[block, minimum width=2cm, minimum height=0.5cm, line width=0.3mm, align=center, fill=blue!10] (ff3) at (0,1.0) {Feed \\ Forward Up};
        \node[block, minimum width=1cm, minimum height=0.5cm, line width=0.3mm, align=center, fill=blue!10] (ff2) at (0,0.2) {Feed \\ Forward Down};
        \node[block, draw, minimum width=0.5cm, minimum height=0.2cm, line width=0.3mm, align=center, fill=yellow!10] (normadap1) at (0,-0.5) {Norm};


        \node[block, minimum width=1cm, minimum height=0.5cm, line width=0.3mm, align=center, fill=yellow!10] (add2) at (0,-1.6) {Add \& Norm};
        \node[block, minimum width=1cm, minimum height=0.5cm, line width=0.3mm, align=center, fill=blue!10] (ff1) at (0,-2.5) {Feed \\ Forward};
        \node[block, minimum width=1cm, minimum height=0.5cm, line width=0.3mm, align=center, fill=yellow!10] (add1) at (0,-3.4) {Add \& Norm};
        \node[block, minimum width=1cm, minimum height=0.5cm, line width=0.3mm, align=center, fill=brown!10] (att1) at (0,-4.2) {Multi-Head \\ Attention};

        \draw[->, line width=0.3mm] (-1.1,-2.0) |- (addadap2);
        \draw[->, line width=0.3mm] (-1.1,-2.0) |- (addadap1);

        \draw[line width=0.3mm] (0,-2.0) |- (-1.1,-2.0);
        \draw[line width=0.3mm] (0,-1.25) |- (-1.1,-1.25);

        \draw[->, line width=0.3mm] (-1.5,-3.0) |- (add3);
        
        \draw[->, line width=0.3mm] (-1.5,-3.0) |- (add2);
        \draw[line width=0.3mm] (0,-3.0) |- (-1.5,-3.0);

        \draw[->, line width=0.3mm] (-1.5,-5.0) |- (add1);
        \draw[line width=0.3mm] (0,-5.0) |- (-1.5,-5.0);

        \draw[->, line width=0.3mm] (-0.5,-4.8) -| (0.5, -4.55);
        \draw[->, line width=0.3mm] (-0.5,-4.8) -| (-0.5, -4.55);
        \draw[->, line width=0.3mm] (0,-5.5) -- (-0,-4.55);

        \path[every node/.style={font=\sffamily\small}]
            (att1) edge[line width=0.3mm] node [right] {} (add1)
            (add1) edge[line width=0.3mm] node [right] {} (ff1)
            (ff1) edge[line width=0.3mm] node [right] {} (add2)
            (add2) edge[line width=0.3mm] node [right] {} (normadap1)
            (normadap1) edge[line width=0.3mm] node [right] {} (ff2)
            (ff2) edge[line width=0.3mm] node [right] {} (ff3)
            (ff3) edge[line width=0.3mm] node [right] {} (addadap1)
            (addadap1) edge[line width=0.3mm] node [right] {} (normadap2)
            (normadap2) edge[line width=0.3mm] node [right] {} (addadap2)
            (addadap2) edge[line width=0.3mm] node [right] {} (add3);


        \node[font=\scriptsize\ttfamily\bfseries, text centered, text width=1cm] at (0.7,-0.9) {Adapter};

    \end{tikzpicture}
    \caption{For the individual adapter predictions, we use a Transformer based model with adapter layers inserted after the feed forward layer of the Transformer as proposed by \citet{pfeiffer-etal-2021-adapterfusion}.}
    \label{fig:adapter}
    \end{figure}

\begin{table*}[t]
\scriptsize
    \centering
    \begin{tabular}[t]{cllr}
        \toprule
         &\textbf{Adapter} & \textbf{Description} & \textbf{Weight} \\
         & \\[\dimexpr-\normalbaselineskip+2pt]
        \hline
         \multirow{7}{*}{\rotatebox[origin=c]{90}{Rationality}\hspace{-10mm}}
         & \\[\dimexpr-\normalbaselineskip+2pt]
         &\textbf{Relevance}& \textbf{Does the comment have a relevance for the discussed topic?} &\textbf{0.20908452}\\
         &\textbf{Fact}& \textbf{Is there at least one fact claiming statement in the comment?} &\textbf{0.18285757}\\
         &\textbf{Opinion} & \textbf{Is there a subjective statement made in the comment?} & \textbf{-0.11069402}\\
         &\textbf{Justification} & \textbf{Is at least one statement justified in the comment?} &  \textbf{0.29000763}\\
         &\textbf{Solution Proposals}& \textbf{Does the comment contain a proposal how an issue could be solved?}& \textbf{0.39535126}\\
         &\textbf{Additional Knowledge}&\textbf{Does the comment contain additional knowledge?}& \textbf{0.14655912}\\
         &Question&Does the comment include a true, i.e., non-rhetoric question? & -0.07331445\\
         & \\[\dimexpr-\normalbaselineskip+2pt]
         \hline
         \multirow{5}{*}{\rotatebox[origin=c]{90}{Reciprocity}\hspace{-10mm}}
         & \\[\dimexpr-\normalbaselineskip+2pt]
         &Referencing Users& Does the comment refer to at least one other user or to all users in the community?& -0.03768367\\
         &Referencing Medium&Does the comment refer to the medium, the editorial team or the moderation team? & 0.07019062\\
         &Referencing Contents &Does the comment refer to content, arguments or positions in other comments? & -0.02847408 \\
         &\textbf{Referencing Personal}&\textbf{Does the comment refer to the person or personal characteristics of other users?}&\textbf{0.21126469}\\       
         &Referencing Format & Does the comment refer to the tone, language, spelling or other formal criteria other comments? &  -0.02674237\\
         & \\[\dimexpr-\normalbaselineskip+2pt]
         \hline
         \multirow{7}{*}{\rotatebox[origin=c]{90}{Civility}\hspace{-10mm}}
         & \\[\dimexpr-\normalbaselineskip+2pt]
         &Polite form of Address & Does the comment contain welcome or farewell phrases? & 0.01482095 \\
         &Respect&Does the comment contain expressions of respect or thankfulness? & 0.00732909 \\
         &Screaming&Does the comment contain clusters of punctuation or capitalization intended to imply screaming?& -0.01900971\\
         &Vulgar& Does the comment contain language that is inappropriate for civil discourse?&-0.04995486 \\
         &Insult& Does the comment contain insults towards one or more people? & -0.05884586\\
         &\textbf{Sarcasm}&\textbf{Does the comment contain biting mockery aimed at devaluing the reference object?} & \textbf{-0.15170863}\\
         &Discrimination&Does the comment explicitly or implicitly contain unfair treatment of groups or individuals?& 0.02934227\\
        & \\[\dimexpr-\normalbaselineskip+2pt]
        \hline
        & \\[\dimexpr-\normalbaselineskip+2pt]
         &\textbf{Storytelling}&\textbf{Does the commenter include personal stories or personal experiences?} &\textbf{0.10628146}\\
\bottomrule
    \end{tabular}
    \caption{Correlation weights $w_k$ of all 20 trained deliberative quality adapters. The weights are calculated as the correlation coefficients between the experts' annotations and non-experts' ones.
    The most important indicators for a high quality comment are marked in bold.
    Note that positive correlations correspond to a positive trait in a high quality comment, while negative correlations correspond to negative traits.}
    \label{tab:all_adapters}
\end{table*}

\subsubsection{KODIE \& \#meinfernsehen - the Experts' View}
\label{par:kodie}

The KODIE annotation framework \citep{heinbach2022}, assigns 23 score-based deliberative and further labels on other aspects to each comment.
These annotations were conducted by trained coders with a scientific background, focusing on deliberative criteria such as fact claims, relevance to the discussion topic, and respectful engagement with other users. The deliberative criteria can each be assigned to one of the three main dimensions of deliberation \citep{bachtiger2009measuring,esau2021different,10.1007/978-3-642-15158-3_3,10.1111/jcom.12104,doi:10.1177/1461444804041444}: 
\begin{description}
\item\textbf{Rationality}, measured by indicators such as reasoning, solution proposals, and provision of additional knowledge.
\item\textbf{Reciprocity}, measured as mutual references between users within a discussion.
\item\textbf{Civility}, measured as the presence of a respectful interaction with others and the absence of insults, pejorative speech, and other markers of disrespect.
\end{description}
The following coding scheme was used:  all categories were coded on a four-point scale from ``clearly not present'' to ``clearly present''. 
Intercoder reliability was tested on a subset of 130 comments and exceeded the critical threshold of Krippendorff’s $\alpha$ of .67 for all categories (\O~=~.83).  The \#meinfernsehen data is annotated with the same scheme as KODIE. For \#meinfernsehen intercoder reliability was tested on 159 comments, exceeding the critical threshold of Krippendorff’s $\alpha$ of .67 for 20 out of 21 categories (\O~=~.74).

We selected 19 out of the 23 deliberative quality criteria to train adapters, since some annotated aspects, e.g., \emph{threat of violence} were not found in the data. In addition to the deliberative quality criteria, we included \emph{storytelling}, which is considered a type II deliberation criterion, according to \citet{Bachtiger2009-BCHDDI-2}, since the description of personal experience when suggesting a solution contributes to the perceived quality of a comment \citep{falk-lapesa-2023-storyarg}. The 20 deliberative aspects that we use are listed in Table~\ref{tab:all_adapters}. After filtering out data points with missing annotations and coding errors, we were left with a total of 13,069 comments to train our adapter models. In the following we will write 
\begin{equation}
    s_k(i)\in \{0, 1, 2, 3\}
\end{equation}
for the $k$-th score ($1\le k\le 20$) of the $i$-th comment ($1\le i\le 13,069$).

\subsubsection{CrowdAnno - the Non-Experts' View}
In the CrowdAnno project, \citet{wilms2023} gathered data on non-experts' perception of uncivil, deliberative, and fact-claiming communication within German online comments through crowd annotation. The dataset includes 13,677 comments from different news media comment sections and online citizen participation projects, annotated by 681 crowdworkers. For AQuA, we used a subset of 1,742 comments that are identical to the KODIE and \#meinfernsehen data. Crowd workers were tasked with evaluating, whether a comment is perceived as enriching and value-adding to the discussion or not, i.e., marking if it contains enriching communication, which could serve as a proxy for deliberative quality. The final score is aggregated from evaluations by 9 different crowd annotators via majority vote. To minimize annotator bias, the crowd workers were sampled to reflect various sociodemographic and educational backgrounds. We will write 
\begin{equation}
    c(i)\in\{0,1\}
\end{equation}
for the binary deliberativeness label of the $i$-th comment.

\subsection{Training the Adapters}
\label{subsec:adaptertrain}
To automatically predict the various deliberation criteria, we use pre-trained language models, such as BERT \citep{devlin-etal-2019-bert}.  We follow the adapter approach:  adapters are extra weights $\theta_k$, that are plugged into pre-trained language models and then learned for a specific task $k$.  The adapted language model for the $k$-th deliberation criterion is written as $f_{\theta_k}(x)$, where $x$ is some text input.  Note that while learning these extra weights, we do not alter the pre-trained model weights.  
More precisely, we used the adapter architecture proposed by \citet{pfeiffer-etal-2021-adapterfusion}, which is shown in Figure~\ref{fig:adapter}. We trained 20 individual adapters to predict scores $f_{\theta_k}(x)$ for individual indicators for deliberative quality in user comments for the KODIE dataset. For training we perform a 65\% (train), 15\% (val), 20\% (test) split on our dataset, resulting in 8,495 training data points, 1,960 for validation and 2,614 for testing.
 Each of the 20 adapters for AQuA is trained with a multi-label classification objective, minimizing the cross entropy loss. We train each adapter for 10 epochs and save the model with the best macro F1 score. 

\subsection{Calculating the Weights}

Assigning an importance to the individual quality dimensions for the overall quality measurement is not a simple task.   Our intuition for weighting the deliberative criteria is to include the perception of people who potentially read and write these comments. 
 For that reason we linked the scientific theory of deliberation to the view of non-scientists by combining the datasets described in detail in Section~\ref{subsec:datasets}.
More precisely, we obtain the weight for each deliberative criterion $k$ by calculating the correlation coefficient,
\begin{equation}
  w_k = \frac{ \sum_{i=1}^{N}(s_k(i)-\bar{s}_k)(c(i)-\bar{c}) }{%
        \sqrt{\sum_{i=1}^{N}(s_k(i)-\bar{s}_k)^2}\sqrt{\sum_{i=1}^{N}(c(i)-\bar{c})^2}},
\end{equation}
between the scientific label $s_k(i)$ (with mean $\bar{s}_k$) for each of the $K=20$ aspects of deliberation and the perception of crowd workers on the comments deliberativeness $c(i)$ (with mean $\bar{c}$) for all $N$ comments.  Note that $w_k$ is a value from the interval between $-1$ and $1$.

\subsection{Building the AQuA Score}

We build an overall quality score $s(x)$ for each comment as the weighted sum of the weights $w_k$ and the predicted score $f_{\theta_k}(x)$ for each of the $K=20$ quality adapters:
\begin{equation}
   s(x) = \sum_{k=1}^K w_k f_{\theta_k}(x).
\end{equation}
The highest and lowest possible scores depend on the number $K$ of criteria and on the range of the predictions $f_{\theta_k}(x)$.  Since the labels from KODIE are from the set $\{0, 1, 2, 3\}$, the predictions are also from this set.  The highest possible score can be reached by setting all positively weighted criteria to their maximum value (i.e, 3) and all negatively weighted criteria to their minimum value (i.e, 0),
\begin{equation}
   s_\text{max} = \sum_{k=0}^K 3\cdot w_k\cdot [w_k\ge 0] \approx 4.9893,
\end{equation}
where $[w_k\ge 0]=1$ if $w_k\ge 0$ and zero otherwise.  Similarly,  the smallest possible score is 
\begin{equation}
    s_\text{min} = \sum_{k=0}^K 3\cdot w_k\cdot [w_k\le 0] \approx -1.6693.
\end{equation}
To get a more intuitive range of values, we scale $s(x)$ to an interval between 0 and 5:
\begin{equation}
 s_\text{AQuA}(x) = 5\cdot \frac{(s(x) - s_\text{min})}{(s_\text{max}-s_\text{min})},
\end{equation}
which is the definition of our proposed AQuA score.  Figure~\ref{fig:aqua} graphically illustrates, how the AQuA score is calculated for a given input comment. 

\subsection{Applying the Score to English Comments}
To apply our method to English datasets, we used the \texttt{wmt19-en-de-model}\footnote{\url{https://huggingface.co/facebook/wmt19-en-de}} \citep{ng-etal-2019-facebook}, to automatically translate all comments in the examined dataset from English to German. Another alternative would be to train adapter models on English data. Since the KODIE dataset consists of German Facebook comments on political issues, discussing German politicians as well, we decided not to translate these comments to train adapter models, but to translate English comments and use the pre-trained German models for evaluation. 

\begin{table}
\centering
    \scriptsize
    \begin{tabular}{clccc}
        \toprule
         &&\textbf{German}&\multicolumn{2}{c}{\textbf{Multilingual BERT}}\\ 
         && \textbf{BERT}&\textbf{cased} & \textbf{uncased}\\
         & \\[\dimexpr-\normalbaselineskip+2pt]
        \hline
         \multirow{7}{*}{\rotatebox[origin=c]{90}{Rationality}\hspace{-3mm}}
         & \\[\dimexpr-\normalbaselineskip+2pt]
         &Relevance&\textbf{0.39}&0.37&0.37\\
         &Fact&\textbf{0.58}&0.56&0.54\\
         &Opinion&\textbf{0.59}&0.57&0.5\\
         &Justification&\textbf{0.7}&0.69&0.67\\
         &Solution Proposals&0.77&\textbf{0.79}&0.76\\
         &Additional Knowledge&0.71&\textbf{0.78}&0.74\\
         &Question&0.84&\textbf{0.87}&\textbf{0.87}\\
         & \\[\dimexpr-\normalbaselineskip+2pt]
         \hline
         \multirow{5}{*}{\rotatebox[origin=c]{90}{Reciprocity}\hspace{-3mm}}
         & \\[\dimexpr-\normalbaselineskip+2pt]
         &Referencing Users&0.86&\textbf{0.88}&0.87\\
         &Referencing Medium&0.92&0.93&\textbf{0.94}\\
         &Referencing Contents&0.7&\textbf{0.81}&0.8\\
         &Referencing Personal& 0.83&\textbf{0.92}&\textbf{0.92}\\
         &Referencing Format&0.89&\textbf{0.96}&\textbf{0.96}\\
         & \\[\dimexpr-\normalbaselineskip+2pt]
         \hline
         \multirow{7}{*}{\rotatebox[origin=c]{90}{Civility}\hspace{-3mm}}
         & \\[\dimexpr-\normalbaselineskip+2pt]
         &Polite form of Address &0.96 & 0.97&\textbf{0.98}\\
         &Respect&0.81&0.9&\textbf{0.91}\\
         &Screaming&0.77&\textbf{0.81}&0.79\\
         &Vulgar&0.76&0.74&\textbf{0.86}\\
         &Insults&0.87&0.87&0.87\\
         &Sarcasm&\textbf{0.48}&\textbf{0.48}&0.34\\
         &Discrimination&0.83&\textbf{0.88}&0.87\\
        \hline
        & \\[\dimexpr-\normalbaselineskip+2pt]
         &Storytelling&0.83&0.85&\textbf{0.86}\\
         & \\[\dimexpr-\normalbaselineskip+2pt]
         \hline 
        & \\[\dimexpr-\normalbaselineskip+2pt]
        &\O \, Total Average (F1-Score) &0.7545&\textbf{0.7815}&0.771\\
        \bottomrule
    \end{tabular}
    \caption{Base models. We analyze the performance of different base models with adapter training on the 20 deliberative aspects. We show the weighted average F1 score.
    Overall, the multilingual BERT cased model performs best on the KODIE test dataset. We therefore use multilingual BERT as a base model for the AQuA score.}
    \label{tab:basemodels}
\end{table}

\section{Analysis and Experiments}

After defining the AQuA score in the previous sections, we briefly discuss the choice of our base model and then analyze the weights that we calculated for the individual adapter predictions.   Finally, we conduct several experiments to show that our model can successfully predict deliberative quality in user comments.

\subsection{Choice of the Base Model}
The correlation coefficients are one important part that affect the composition of AQuA. The other part are the predictions of each of the 20 trained adapters. The adapter weights can be trained with different base architectures. 
To determine which base model performs best, we examine the performance of different models, namely German BERT Base cased~\citep{chan-etal-2020-germans} and multilingual BERT~\citep{devlin-etal-2019-bert} in the cased and uncased variants, on the KODIE test split. The training procedure is the same as described in Section~\ref{subsec:adaptertrain}. The results are shown in Table~\ref{tab:basemodels}. As the datasets are highly imbalanced, and some deliberative qualities do not occur often in the training data, we report the weighted averaged F1 score, i.e., a global weighted average F1 score for each class.
The trained adapter weights with the multilingual BERT model as base model outperform the German BERT model on 15 out of the 20 tasks. In direct comparison, the cased variant of Multilingual BERT performs slightly better than the uncased one. Based on these results we take the multilingual BERT Base cased model\footnote{\url{https://huggingface.co/bert-base-multilingual-cased}} as our base model for calculating the AQuA score.

\begin{table}[tbp]
    \centering
    \scriptsize
    \begin{tabular}{clcccc}
    \toprule
     &\textbf{Label} & \multicolumn{4}{c}{\textbf{Frequency}}  \\
     & & 0 & 1 & 2 & 3 \\
     \hline
        \multirow{7}{*}{\rotatebox[origin=c]{90}{Rationality}\hspace{-3mm}}
         & \\[\dimexpr-\normalbaselineskip+2pt]
         &Relevance&130&200&345&1065\\
         &Fact&1155&113&155&317\\
         &Opinion&27&15&13&123\\
         &Justification&1177&78&139&346\\
         &Solution Proposals&932&400&281&127\\
         &Additional Knowledge&1524&76&91&48\\
         &Question&1590&55&45&50\\
         & \\[\dimexpr-\normalbaselineskip+2pt]
         \hline
         \multirow{5}{*}{\rotatebox[origin=c]{90}{Reciprocity}\hspace{-3mm}}
         & \\[\dimexpr-\normalbaselineskip+2pt]
         &Referencing Users&1164&128&62&386\\
         &Referencing Medium&173&1&1&3\\
         &Referencing Contents&1142&98&119&381\\
         &Referencing Personal&177&1&0&0\\
         &Referencing Format&177&0&0&1\\
         & \\[\dimexpr-\normalbaselineskip+2pt]
         \hline
         \multirow{7}{*}{\rotatebox[origin=c]{90}{Civility}\hspace{-3mm}}
         & \\[\dimexpr-\normalbaselineskip+2pt]
         &Polite form of Address &1725&3&6&6\\
         &Respect&1572&25&100&43\\
         &Screaming&1612&30&53&45\\
         &Vulgar&1654&44&23&19\\
         &Insults&1670&29&21&20\\
         &Sarcasm&1327&115&130&168\\
         &Discrimination&170&2&1&5\\
         & \\[\dimexpr-\normalbaselineskip+2pt]
        \hline
        & \\[\dimexpr-\normalbaselineskip+2pt]
         &Storytelling&1617&59&46&18\\
         & \\[\dimexpr-\normalbaselineskip+2pt]
         \hline
    \bottomrule
    \end{tabular}
    \caption{CrowdAnno. Absolute frequencies of each label in the subset of the CrowAnno dataset, used to calculate the correlation coefficients.}
    \label{tab:frequency}
\end{table}

\subsection{Insights from the Correlations}
The calculated correlation coefficients serve as weights in AQuA to give more importance to some deliberative aspects than others. Besides their values determining the importance for each criterion, the sign of the correlation coefficient reveals if an aspect is positively or negatively associated with comment quality. In the following, we discuss the coefficients and examine whether findings from previous deliberative research are consistent with our results. The coefficients with large absolute values are marked bold in Table~\ref{tab:all_adapters}. 

For an overview of the data distribution, Table \ref{tab:frequency} lists the absolute frequencies of each label for each deliberative quality criteria in the subset of the KODIE and \#meinfernsehen datasets that have been annotated using the CrowdAnno framework. These points were used to calculate the correlation coefficients. Note that these are not the frequencies in the dataset used for training the adapters. However, the small subset reflects the class imbalance that is present in the data, indicating that some categories such as vulgar language, insults and even storytelling do not occur often. 

It is striking that nearly all indicators for \emph{rationality} are strongly positively correlated with non-experts' perceived deliberative quality of comments. Using well-reasoned arguments that are relevant to the topic has been found to be an important aspect in distinguishing between comments of high and low deliberative quality \citep{diakopoulos2015picking,kolhatkar2020sfu}. Unfounded expressions of opinion, on the other hand, are perceived as non-constructive, i.e., negative, in user comments. Our results support that finding, as opinion is highly negatively correlated with the perceived deliberative quality.

Of all the indicators of \emph{reciprocity}, referring to personal characteristics of others has the greatest positive impact on the overall score. This is surprising as deliberative literature primarily highlights engaging with others' positions, not their personal traits, as a quality indicator \citep[e.g.,][]{doi:10.1177/0093650218797884}. 

Within the \emph{civility} criteria, sarcasm stands out with a rather high negative correlation coefficient. Sarcasm, as well as doubting, criticism, and insults have been identified as one form of expressing disrespect towards other participants \citep{10.5555/2021109.2021116}. 
The large correlation weight for sarcasm is a stable finding, since it is more frequent in the KODIE data, in contrast to insults. 

While not being a central aspect of deliberation, storytelling in form of personal anecdotes can foster empathy and mutual understanding between participants and resolve differences \citep{black2008listening}. Thus, it is reasonable that \emph{storytelling} plays an important role in the weighting of AQuA, as well.

\subsection{Evaluating the Score}

Having trained the AQuA score using the KODIE, \#meinfernsehen and CrowdAnno datasets, we next show that the learned adapter weights and correlations transfer to other datasets as well and give scores that are qualitatively and also quantitatively convincing.

\begin{table}[tbp]
    \centering
    \scriptsize
    \begin{tabular}{clc}
    \toprule
     &\textbf{Adapter} & \textbf{F1 Score}  \\ \\
     \hline
        \multirow{7}{*}{\rotatebox[origin=c]{90}{Rationality}\hspace{-3mm}}
         & \\[\dimexpr-\normalbaselineskip+2pt]
         &Relevance&13.22\\
         &Fact&18.48\\
         &Opinion&42.93\\
         &Justification&29.49\\
         &Solution Proposals&56.04\\
         &Additional Knowledge&38.97\\
         &Question&62.25\\
         & \\[\dimexpr-\normalbaselineskip+2pt]
         \hline
         \multirow{5}{*}{\rotatebox[origin=c]{90}{Reciprocity}\hspace{-3mm}}
         & \\[\dimexpr-\normalbaselineskip+2pt]
         &Referencing Users&66.85\\
         &Referencing Medium&69.23\\
         &Referencing Contents&66.28\\
         &Referencing Personal&70.40\\
         &Referencing Format&70.40\\
         & \\[\dimexpr-\normalbaselineskip+2pt]
         \hline
         \multirow{7}{*}{\rotatebox[origin=c]{90}{Civility}\hspace{-3mm}}
         & \\[\dimexpr-\normalbaselineskip+2pt]
         &Polite form of Address &69.89\\
         &Respect&69.67\\
         &Screaming&67.96\\
         &Vulgar&65.64\\
         &Insults&70.40\\
         &Sarcasm&66.12\\
         &Discrimination&65.84\\
         & \\[\dimexpr-\normalbaselineskip+2pt]
        \hline
        & \\[\dimexpr-\normalbaselineskip+2pt]
         &Storytelling&65.33\\
         & \\[\dimexpr-\normalbaselineskip+2pt]
         \hline
    \bottomrule
    \end{tabular}
    \caption{SOCC. Adapters that align with toxicity reach a high weighted average F1 score with toxicity levels from the SOCC dataset.}
    \label{tab:sfu}
\end{table}

\begin{figure}[htbp]
	\centering
	\begin{tikzpicture}
		\node[draw, thin, color=white, fill=white, rounded corners=0.0mm, inner sep=1mm] (img1) {
			\includegraphics[scale=0.51]{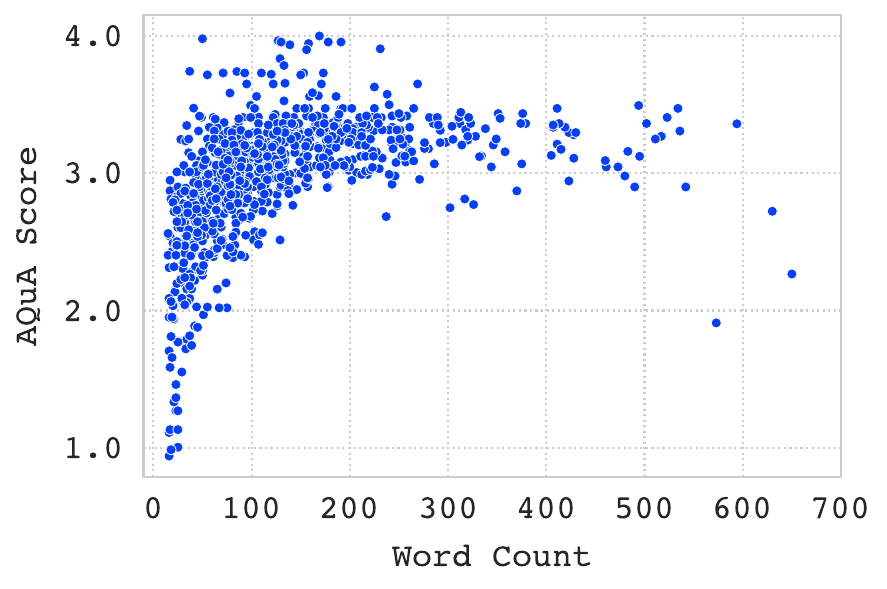}
		};
		\node[text width=6cm, text centered, above of=img1] at (0.3, 1.7) {\fontfamily{pcr}\selectfont\scriptsize{AQuA Scores Europolis}};	
	\end{tikzpicture}

    \caption{Europolis. AQuA scores (y-axis) vs the comment length (x-axis, word count) rule out that comment length alone is a factor for a high AQuA score. }
    \label{fig:length}
\end{figure}

\begin{table*}
    \centering
    \scriptsize
    \begin{tabularx}{\textwidth}{ >{\hsize=0.65\hsize\RaggedRight}X 
                                  >{\hsize=0.15\hsize\RaggedRight}X
                                  >{\hsize=0.15\hsize\RaggedRight}X
                                  >{\hsize=0.05\hsize\RaggedRight}X}
        \toprule
        & \\[\dimexpr-\normalbaselineskip+2pt]
        \multicolumn{4}{c}{\scriptsize{\textbf{Top 3 Comments from Europolis}}}\\
        \textbf{Comment} &\textbf{Europolis Labels} &\textbf{Adapter Predictions}&\textbf{Score} \\
        & \\[\dimexpr-\normalbaselineskip+2pt]
        \hline
        \vspace{0.2em}
        The problem with the whole story is that first of all the cost of living has to be equalized - that includes, of course, wages, or salaries. If that - I assume we are only Poles and Germans here - and an Austrian, excuse me Julian - that we, I think, as I have come to know it - I have just said, we have a twin town in Poland - the cost of living was at least two years ago in Poland much lower than in Germany and then of course higher wages have to be paid here, so that you can buy the piece of bread, which is correspondingly lower in Poland and that's why Frankfurt/Oder to the other side is a constant border traffic. Buying gas in Poland is just much cheaper than in Frankfurt/Oder on the border. So the problem is simply that the cost of living in the individual states is so different that you can't equate it with wages and salaries at all.&
        \vspace{0.2em}
        \begin{minipage}[t]{\linewidth}\emph{interact.:~2,\\ respect:~1,\\ storytelling:~1,\\ justification:~2}\end{minipage}&
        \vspace{0.2em}
        \begin{minipage}[t]{\linewidth}\emph{rel.:~3, \ fact:~3,\\ opinion:~3,\\ justification:~3,\\ suggest. sol.:~3,\\ additional know.:~3,\\ storytelling:~3}\end{minipage} & 
        \vspace{0.2em}
        4.0005\\
        \multicolumn{4}{c}{} \\ [0.5em] 
        \hline
        \vspace{0.2em}
        Financial problems always existed in different countries. If someone wants to live in another country, he can always do so. So if he/she wants to work a few years in some country in order to send the family money that he/she earned, he/she should not be prevented from doing so.& 
        \vspace{0.2em}
        \begin{minipage}[t]{\linewidth}\emph{interact.:~3,\\ respect:~1,\\ justification:~3,\\ common good:~2}\end{minipage}&
        \vspace{0.2em}
        \begin{minipage}[t]{\linewidth}\emph{rel.:~3, \ fact:~3,\\ justification:~3,\\ suggest. sol.:~3,\\ additional know.:~2,\\ storytelling:~1}\end{minipage} &
        \vspace{0.2em}
        3.9803\\
        \multicolumn{4}{c}{} \\ [0.5em] 
        \hline
        \vspace{0.2em}
        Many people are coming to other countries not just because of economic reasons. Often, they are persecuted in their own countries on the religious grounds and they are trying to find asylum in another country. Then, the government should give them political asylum, papers or right of permanent residency and then they can work. For example Germany is rich enough to give jobs for immigrants and integrate them in the society because the society is aging and somebody has to work for the new generation which would like to get future pensions or something like that. Society is aging so they need immigrants. Similar to Poland where the government should legalize immigrants in a similar way. It is hard to say how it actually should look like.& 
        \vspace{0.2em}
        \begin{minipage}[t]{\linewidth}\emph{respect:~2,\\ justification:~2,\\ common good:~1}\end{minipage}&
        \vspace{0.2em}
        \begin{minipage}[t]{\linewidth}\emph{rel.:~3, \ fact:~3,\\ suggest. sol.:~3,\\ additional know.:~2.,\\ justification:~3,\\ discrim.:~3}\end{minipage} &
        \vspace{0.2em}
        3.9666\\
        \multicolumn{4}{c}{} \\ [2em] 
        \midrule
         & \\[\dimexpr-\normalbaselineskip+2pt]
        \multicolumn{4}{c}{\scriptsize{\textbf{Lowest 3 Comments from Europolis}}}\\
        \textbf{Comment} &\textbf{Europolis Labels} & \textbf{Adapter Predictions}&\textbf{Score} \\
        & \\[\dimexpr-\normalbaselineskip+2pt]
        \hline
        \vspace{0.2em}
        A question for Udo: To what dimension is the problem with the migration of workers growing? &
        \vspace{0.2em}
        \begin{minipage}[t]{\linewidth}\emph{interact.:~2,\\ respect:~1}\end{minipage}&
        \vspace{0.2em}
        \begin{minipage}[t]{\linewidth}\emph{question:~3,\\ref. user:~3,\\ ref. content:~3}\end{minipage}&
        \vspace{0.2em}
        0.9393\\
        \multicolumn{4}{c}{} \\ [0.5em] 
        \hline
        \vspace{0.2em}
        Thank you very much. Aurore, you also wanted to say something especially before the break but now too?&
        \vspace{0.2em}
        \begin{minipage}[t]{\linewidth}\emph{interact.:~2, \\respect:~1,\\ storytelling:~1, \\justification:~2,\\ common good:~2}\end{minipage}&
        \vspace{0.2em}
        \begin{minipage}[t]{\linewidth}\emph{fact:~1, \ question:~3,\\ ref. user:~3, \\ref. content:~3, \\ polite addr.:~2,\\ sarcasm:~1}\end{minipage}&
        \vspace{0.2em}
        0.9849\\
        \multicolumn{4}{c}{} \\ [0.5em] 
        \hline
        \vspace{0.2em}
        To tell you the truth, I do not know what is discussed? Are we talking about the quotas – how many people could come here?& 
        \vspace{0.2em}
        \begin{minipage}[t]{\linewidth}\emph{respect:~1, \\storytelling:~1,\\ justification:~1,\\ common good:~1}\end{minipage}&
        \vspace{0.2em}
        \begin{minipage}[t]{\linewidth}\emph{question:~3, \\ref. user:~3}\end{minipage}&
        \vspace{0.2em}
        1.0034\\

        \bottomrule
    \end{tabularx}
    \caption{Europolis. Top 3 comments with the highest and top 3 comments with the lowest calculated AQuA scores. We only show the scores and the predicted labels of the individual adapters where the prediction is larger than zero.  The original labels (from Europolis, 5 labels) show that the AQuA score is well aligned with the original labels.} 
    \label{tab:evaleuro}
\end{table*}


\subsubsection{SFU Opinion and Comments Corpus}
We predict AQuA scores on comments of the SFU opinion and comment corpus (SOCC) \citep{kolhatkar2020sfu}. The dataset includes 1,121 comments on news articles that have been annotated for \emph{constructiveness} (binary annotations) and \emph{toxicity} (four point scale from not toxic to very toxic). According to \citet{kolhatkar2020sfu}, constructive comments are required ``to create a civil dialogue through remarks that are relevant to the article and not intended to merely provoke an emotional response''.

We calculate AQuA scores and use them to predict the binary constructive label for each comment in the SOCC. Choosing a threshold of 2.3, i.e., inferring $\hat{y}_\text{constructive} = 1$, if $s_\text{AQuA} \ge$ 2.3, we get an F1 score of 81.73. Note that the threshold is a hyperparameter and a value of 2.3 was chosen, because with performed best on the data. As the dataset also comprises labels for toxic comments, we use the individual adapter predictions for \emph{screaming}, \emph{vulgar}, \emph{insults}, \emph{sarcasm}, and \emph{discrimination} to predict the level of toxicity for each comment. Both the SOCC labels $y_\text{toxic}$ as well as our predictions $s_k(i)$ are numbers from 0 to 3, therefore we simply use the individual predictions of each adapter as an indicator for the toxicity level and calculate the weighted average F1 score.
With 829 comments labeled as not toxic at all (label 0), 172 with label 1, 35 with label 2 and only 7 comments that are marked as clearly toxic (label 3), the distribution is very similar to the one we see in the datasets we used for AQuA. Table~\ref{tab:sfu} shows that we reach good F1 scores for adapters that align with toxicity.

\subsubsection{Europolis}
For a qualitative analysis of the AQuA score, we apply it to the Europolis dataset \citep{Gerber_Baechtiger_Shikano_Reber_Rohr_2018}. 
Europolis includes transcribed speech contributions of a deliberative poll on migration and climate change, 
annotated for \emph{interactivity}, \emph{respect}, \emph{storytelling}, \emph{justification} and \emph{common good}. 
We calculate AQuA scores for each contribution in the dataset and report the top 3 highest and lowest ranked comments in Table~\ref{tab:evaleuro}. 
For interpretability, we list both the predicted labels of the individual adapters and the original Europolis labels (in both cases only for values greater than 0).
While both differ, the AQuA labels approximately match the original Europolis labels. The top 3 comments are all rated highly with positive deliberative aspects such as storytelling, justification and additional knowledge, while the lowest comments exhibit negative deliberative aspects such as 
sarcasm and references to other participants. Overall, all of the the lowest scored comments are  questions to clarify certain aspects in the discussion, whereas the higher scored comments consist
of sophisticated opinions. 

When comparing the AQuA predictions to the original Europolis labels, we find that the AQuA score  seems consistent with the original labels, while enhancing the prediction since the AQuA score consists of 20 deliberative aspects instead of the 5. This demonstrates the value of AQuA as a unified score that can be applied to any dataset based on the chosen deliberative aspects.


\paragraph{Does comment length matter?}
An interesting observation is that the lowest ranked comments in the dataset are much shorter than the high ranked ones.   To study whether comment length alone is the most important factor that causes our model to predict a large score, we take a closer look at the distribution of scores depending on the length of the comment. Figure~\ref{fig:length} displays the AQuA score (y-axis) in comparison to the comment length (x-axis, word count).  While it is true that short comments get the lowest scores, which is probably due to the fact that they do not have much content, the visual analysis reveals also that medium length comments get the highest scores.  This rules out that comment length is the most relevant factor for our score. 




\section{Conclusion}
In this work we introduce AQuA, an approach for an automated deliberative quality score based on large language models and adapters. The score combines annotations of experts and the view of non-experts on real online discussion comments. 

We show that the trained adapters are capable of predicting individual scores for different aspects of deliberative quality and that the overall score aggregates these predictions in a meaningful way. The correlation coefficients between experts' and non-experts' annotations reveal the most important positive and negative deliberative aspects, which allows us to confirm theoretical and empirical findings in deliberation literature into AQuA.

Furthermore, we evaluate our score (trained on KODIE and CrowdAnno) on two further datasets (SOCC and Europolis) to show that the predictions of the learned adapters transfer well to unseen datasets. First, we show that the adapter predictions that build the AQuA score are useful for classifying constructive and toxic comments on the SOCC dataset.  Then we perform a qualitative analysis of the AQuA score by manual assessing the top 3 and bottom 3 scored comments in the Europolis dataset and show that comments with well formed opinions receive large scores, while comments providing little value to the discussion receive lower scores.

Overall, we show that AQuA can be used successfully to automatically assess deliberative quality while aligning with theoretical and empirical background in deliberation literature.

\newpage
\section{Bibliographical References}\label{sec:reference}

\bibliographystyle{lrec-coling2024-natbib}
\bibliography{aqua-lrec-coling2024}

\begin{thebibliography}{44}
\expandafter\ifx\csname natexlab\endcsname\relax\def\natexlab#1{#1}\fi

\bibitem[{Acheampong et~al.(2020)Acheampong, Wenyu, and
  Nunoo-Mensah}]{https://doi.org/10.1002/eng2.12189}
Francisca~Adoma Acheampong, Chen Wenyu, and Henry Nunoo-Mensah. 2020.
\newblock \href {https://doi.org/https://doi.org/10.1002/eng2.12189}
  {Text-based emotion detection: Advances, challenges, and opportunities}.
\newblock \emph{Engineering Reports}, 2(7):e12189.

\bibitem[{B\"{a}chtiger et~al.(2009)B\"{a}chtiger, Niemeyer, Neblo,
  Steenbergen, and Steiner}]{Bachtiger2009-BCHDDI-2}
Andr\'e B\"{a}chtiger, Simon Niemeyer, Michael Neblo, Marco~R. Steenbergen, and
  J\"{u}rg Steiner. 2009.
\newblock \href {https://doi.org/10.1111/j.1467-9760.2009.00342.x}
  {Disentangling diversity in deliberative democracy: Competing theories, their
  blind spots and complementarities}.
\newblock \emph{Journal of Political Philosophy}, 18(1):32--63.

\bibitem[{B{\"a}chtiger et~al.(2009)B{\"a}chtiger, Shikano, Pedrini, and
  Ryser}]{bachtiger2009measuring}
Andr{\'e} B{\"a}chtiger, Susumu Shikano, Seraina Pedrini, and Mirjam Ryser.
  2009.
\newblock \href {https://ash.harvard.edu/files/ash/files/baechtiger_0.pdf}
  {Measuring deliberation 2.0: standards, discourse types, and
  sequenzialization}.
\newblock In \emph{ECPR General Conference}, pages 5--12. Potsdam.

\bibitem[{Beauchamp(2020)}]{10.1093/oxfordhb/9780190460518.013.23}
Nick Beauchamp. 2020.
\newblock \href {https://doi.org/10.1093/oxfordhb/9780190460518.013.23}
  {{321Modeling and Measuring Deliberation Online}}.
\newblock In \emph{{The Oxford Handbook of Networked Communication}}. Oxford
  University Press.

\bibitem[{Bender et~al.(2011)Bender, Morgan, Oxley, Zachry, Hutchinson, Marin,
  Zhang, and Ostendorf}]{10.5555/2021109.2021116}
Emily~M. Bender, Jonathan~T. Morgan, Meghan Oxley, Mark Zachry, Brian
  Hutchinson, Alex Marin, Bin Zhang, and Mari Ostendorf. 2011.
\newblock \href {https://aclanthology.org/W11-0707.pdf} {Annotating social
  acts: authority claims and alignment moves in wikipedia talk pages}.
\newblock In \emph{Proceedings of the Workshop on Languages in Social Media},
  LSM '11, page 48–57, USA. Association for Computational Linguistics.

\bibitem[{Black(2008)}]{black2008listening}
Laura~W Black. 2008.
\newblock \href {https://doi.org/https://doi.org/10.16997/jdd.76} {Listening to
  the city: Difference, identity, and storytelling in online deliberative
  groups}.
\newblock \emph{Journal of Deliberative Democracy}, 5(1).

\bibitem[{Bächtiger et~al.(2022)Bächtiger, Gerber, and
  Fournier-Tombs}]{10.1093/oso/9780192848925.003.0006}
André Bächtiger, Marlène Gerber, and Eléonore Fournier-Tombs. 2022.
\newblock \href {https://doi.org/10.1093/oso/9780192848925.003.0006}
  {{83Discourse Quality Index}}.
\newblock In \emph{{Research Methods in Deliberative Democracy}}. Oxford
  University Press.

\bibitem[{Chan et~al.(2020)Chan, Schweter, and
  M{\"o}ller}]{chan-etal-2020-germans}
Branden Chan, Stefan Schweter, and Timo M{\"o}ller. 2020.
\newblock \href {https://doi.org/10.18653/v1/2020.coling-main.598}
  {{G}erman{'}s next language model}.
\newblock In \emph{Proceedings of the 28th International Conference on
  Computational Linguistics}, pages 6788--6796, Barcelona, Spain (Online).
  International Committee on Computational Linguistics.

\bibitem[{Chen(2017)}]{chen2017online}
Gina~Masullo Chen. 2017.
\newblock \emph{Online incivility and public debate: Nasty talk}.
\newblock Springer.

\bibitem[{Coe et~al.(2014)Coe, Kenski, and Rains}]{10.1111/jcom.12104}
Kevin Coe, Kate Kenski, and Stephen~A. Rains. 2014.
\newblock \href {https://doi.org/10.1111/jcom.12104} {{Online and Uncivil?
  Patterns and Determinants of Incivility in Newspaper Website Comments}}.
\newblock \emph{Journal of Communication}, 64(4):658--679.

\bibitem[{Devlin et~al.(2019)Devlin, Chang, Lee, and
  Toutanova}]{devlin-etal-2019-bert}
Jacob Devlin, Ming-Wei Chang, Kenton Lee, and Kristina Toutanova. 2019.
\newblock \href {https://doi.org/10.18653/v1/N19-1423} {{BERT}: Pre-training of
  deep bidirectional transformers for language understanding}.
\newblock In \emph{Proceedings of the 2019 Conference of the North {A}merican
  Chapter of the Association for Computational Linguistics: Human Language
  Technologies, Volume 1 (Long and Short Papers)}, pages 4171--4186,
  Minneapolis, Minnesota. Association for Computational Linguistics.

\bibitem[{Diakopoulos(2015)}]{diakopoulos2015picking}
Nicholas Diakopoulos. 2015.
\newblock \href
  {https://isojjournal.wordpress.com/2015/04/15/picking-the-nyt-picks-editorial-criteria-and-automation-in-the-curation-of-online-news-comments/}
  {Picking the nyt picks: Editorial criteria and automation in the curation of
  online news comments}.
\newblock \emph{ISOJ Journal}, 5(1):147--166.

\bibitem[{Dryzek(2002)}]{dryzek2002deliberative}
John~S Dryzek. 2002.
\newblock \emph{Deliberative democracy and beyond: Liberals, critics,
  contestations}.
\newblock Oxford University Press, USA.

\bibitem[{Esau et~al.(2021)Esau, Fleu{\ss}, and Nienhaus}]{esau2021different}
Katharina Esau, Dannica Fleu{\ss}, and Sarah-Michelle Nienhaus. 2021.
\newblock \href {https://doi.org/https://doi.org/10.1002/poi3.232} {Different
  arenas, different deliberative quality? using a systemic framework to
  evaluate online deliberation on immigration policy in germany}.
\newblock \emph{Policy \& Internet}, 13(1):86--112.

\bibitem[{Falk and Lapesa(2023{\natexlab{a}})}]{falk-lapesa-2023-bridging}
Neele Falk and Gabriella Lapesa. 2023{\natexlab{a}}.
\newblock \href {https://doi.org/10.18653/v1/2023.findings-eacl.187} {Bridging
  argument quality and deliberative quality annotations with adapters}.
\newblock In \emph{Findings of the Association for Computational Linguistics:
  EACL 2023}, pages 2469--2488, Dubrovnik, Croatia. Association for
  Computational Linguistics.

\bibitem[{Falk and Lapesa(2023{\natexlab{b}})}]{falk-lapesa-2023-storyarg}
Neele Falk and Gabriella Lapesa. 2023{\natexlab{b}}.
\newblock \href {https://doi.org/10.18653/v1/2023.acl-long.132} {{S}tory{ARG}:
  a corpus of narratives and personal experiences in argumentative texts}.
\newblock In \emph{Proceedings of the 61st Annual Meeting of the Association
  for Computational Linguistics (Volume 1: Long Papers)}, pages 2350--2372,
  Toronto, Canada. Association for Computational Linguistics.

\bibitem[{Fournier-Tombs and
  Di~Marzo~Serugendo(2020)}]{fournier-tombs_delibanalysis_2020}
Eleonore Fournier-Tombs and Giovanna Di~Marzo~Serugendo. 2020.
\newblock \href {https://doi.org/10.1177/0165551519871828} {{DelibAnalysis}:
  {Understanding} the quality of online political discourse with machine
  learning}.
\newblock \emph{Journal of Information Science}, 46(6):810--822.

\bibitem[{Friess and Eilders(2015)}]{friess2015systematic}
Dennis Friess and Christiane Eilders. 2015.
\newblock \href {https://doi.org/https://doi.org/10.1002/poi3.95} {A systematic
  review of online deliberation research}.
\newblock \emph{Policy \& Internet}, 7(3):319--339.

\bibitem[{Friess et~al.(2021)Friess, Ziegele, and
  Heinbach}]{doi:10.1080/10584609.2020.1830322}
Dennis Friess, Marc Ziegele, and Dominique Heinbach. 2021.
\newblock \href {https://doi.org/10.1080/10584609.2020.1830322} {Collective
  civic moderation for deliberation? exploring the links between citizens’
  organized engagement in comment sections and the deliberative quality of
  online discussions}.
\newblock \emph{Political Communication}, 38(5):624--646.

\bibitem[{Gerber et~al.(2018)Gerber, Bächtiger, Shikano, Reber, and
  Rohr}]{Gerber_Baechtiger_Shikano_Reber_Rohr_2018}
Marlène Gerber, André Bächtiger, Susumu Shikano, Simon Reber, and Samuel
  Rohr. 2018.
\newblock \href {https://doi.org/10.1017/S0007123416000144} {Deliberative
  abilities and influence in a transnational deliberative poll (europolis)}.
\newblock \emph{British Journal of Political Science}, 48(4):1093–1118.

\bibitem[{Gerlach and Eilders(2022)}]{2022:gerlach:meinferns}
Frauke Gerlach and Christiane Eilders, editors. 2022.
\newblock \href {https://doi.org/https://doi.org/10.5771/9783748928690}
  {\emph{\#meinfernsehen 2021}}.
\newblock Nomos, Baden-Baden.

\bibitem[{Gold et~al.(2015)Gold, El-Assady, Hautli-Janisz, Bögel, Rohrdantz,
  Butt, Holzinger, and Keim}]{10.1093/llc/fqv033}
Valentin Gold, Mennatallah El-Assady, Annette Hautli-Janisz, Tina Bögel,
  Christian Rohrdantz, Miriam Butt, Katharina Holzinger, and Daniel Keim. 2015.
\newblock \href {https://doi.org/10.1093/llc/fqv033} {{Visual linguistic
  analysis of political discussions: Measuring deliberative quality}}.
\newblock \emph{Digital Scholarship in the Humanities}, 32(1):141--158.

\bibitem[{Graham(2010)}]{10.1007/978-3-642-15158-3_3}
Todd Graham. 2010.
\newblock \href {https://doi.org/https://doi.org/10.1007/978-3-642-15158-3_3}
  {The use of expressives in online political talk: Impeding or facilitating
  the normative goals of deliberation?}
\newblock In \emph{Electronic Participation}, pages 26--41, Berlin, Heidelberg.
  Springer Berlin Heidelberg.

\bibitem[{Heinbach et~al.(2022)Heinbach, Wilms, and Ziegele}]{heinbach2022}
Dominique Heinbach, Lena Wilms, and Marc Ziegele. 2022.
\newblock \href
  {https://www.researchgate.net/publication/361260861_Effects_of_empowerment_moderation_in_online_discussions_A_field_experiment_with_four_news_outlets}
  {Effects of empowerment moderation in online discussions: A field experiment
  with four news outlets}.
\newblock In \emph{72nd Annual Conference of the International Communication
  Association (ICA)}.

\bibitem[{Houlsby et~al.(2019)Houlsby, Giurgiu, Jastrzebski, Morrone,
  De~Laroussilhe, Gesmundo, Attariyan, and Gelly}]{pmlr-v97-houlsby19a}
Neil Houlsby, Andrei Giurgiu, Stanislaw Jastrzebski, Bruna Morrone, Quentin
  De~Laroussilhe, Andrea Gesmundo, Mona Attariyan, and Sylvain Gelly. 2019.
\newblock \href {https://proceedings.mlr.press/v97/houlsby19a.html}
  {Parameter-efficient transfer learning for {NLP}}.
\newblock In \emph{Proceedings of the 36th International Conference on Machine
  Learning}, volume~97 of \emph{Proceedings of Machine Learning Research},
  pages 2790--2799. PMLR.

\bibitem[{Kolhatkar et~al.(2020)Kolhatkar, Wu, Cavasso, Francis, Shukla, and
  Taboada}]{kolhatkar2020sfu}
Varada Kolhatkar, Hanhan Wu, Luca Cavasso, Emilie Francis, Kavan Shukla, and
  Maite Taboada. 2020.
\newblock \href {https://doi.org/https://doi.org/10.1007/s41701-019-00065-w}
  {The sfu opinion and comments corpus: A corpus for the analysis of online
  news comments}.
\newblock \emph{Corpus Pragmatics}, 4:155--190.

\bibitem[{Lawrence and Reed(2020)}]{10.1162/coli_a_00364}
John Lawrence and Chris Reed. 2020.
\newblock \href {https://doi.org/10.1162/coli_a_00364} {{Argument Mining: A
  Survey}}.
\newblock \emph{Computational Linguistics}, 45(4):765--818.

\bibitem[{Liu et~al.(2019)Liu, Ott, Goyal, Du, Joshi, Chen, Levy, Lewis,
  Zettlemoyer, and Stoyanov}]{liu2019roberta}
Yinhan Liu, Myle Ott, Naman Goyal, Jingfei Du, Mandar Joshi, Danqi Chen, Omer
  Levy, Mike Lewis, Luke Zettlemoyer, and Veselin Stoyanov. 2019.
\newblock \href {https://doi.org/https://doi.org/10.48550/arXiv.1907.11692}
  {Roberta: A robustly optimized bert pretraining approach}.
\newblock \emph{arXiv preprint arXiv:1907.11692}.

\bibitem[{Mendonca et~al.(2022)Mendonca, Lavie, and
  Trancoso}]{mendonca-etal-2022-qualityadapt}
John Mendonca, Alon Lavie, and Isabel Trancoso. 2022.
\newblock \href {https://doi.org/10.18653/v1/2022.sigdial-1.9}
  {{Q}uality{A}dapt: an automatic dialogue quality estimation framework}.
\newblock In \emph{Proceedings of the 23rd Annual Meeting of the Special
  Interest Group on Discourse and Dialogue}, pages 83--90, Edinburgh, UK.
  Association for Computational Linguistics.

\bibitem[{Ng et~al.(2019)Ng, Yee, Baevski, Ott, Auli, and
  Edunov}]{ng-etal-2019-facebook}
Nathan Ng, Kyra Yee, Alexei Baevski, Myle Ott, Michael Auli, and Sergey Edunov.
  2019.
\newblock \href {https://doi.org/10.18653/v1/W19-5333} {{F}acebook {FAIR}{'}s
  {WMT}19 news translation task submission}.
\newblock In \emph{Proceedings of the Fourth Conference on Machine Translation
  (Volume 2: Shared Task Papers, Day 1)}, pages 314--319, Florence, Italy.
  Association for Computational Linguistics.

\bibitem[{Niemeyer et~al.(2024)Niemeyer, Veri, Dryzek, and
  Bächtier}]{NIEMEYER_VERI_DRYZEK_BAECHTIGER_2024}
Simon Niemeyer, Francesco Veri, John~S. Dryzek, and André Bächtier. 2024.
\newblock \href {https://doi.org/10.1017/S0003055423000023} {How deliberation
  happens: Enabling deliberative reason}.
\newblock \emph{American Political Science Review}, 118(1):345–362.

\bibitem[{Papacharissi(2004)}]{doi:10.1177/1461444804041444}
Zizi Papacharissi. 2004.
\newblock \href {https://doi.org/10.1177/1461444804041444} {Democracy online:
  civility, politeness, and the democratic potential of online political
  discussion groups}.
\newblock \emph{New Media \& Society}, 6(2):259--283.

\bibitem[{Pfeiffer et~al.(2021)Pfeiffer, Kamath, R{\"u}ckl{\'e}, Cho, and
  Gurevych}]{pfeiffer-etal-2021-adapterfusion}
Jonas Pfeiffer, Aishwarya Kamath, Andreas R{\"u}ckl{\'e}, Kyunghyun Cho, and
  Iryna Gurevych. 2021.
\newblock \href {https://doi.org/10.18653/v1/2021.eacl-main.39}
  {{A}dapter{F}usion: Non-destructive task composition for transfer learning}.
\newblock In \emph{Proceedings of the 16th Conference of the European Chapter
  of the Association for Computational Linguistics: Main Volume}, pages
  487--503, Online. Association for Computational Linguistics.

\bibitem[{Pfeiffer et~al.(2020)Pfeiffer, R{\"u}ckl{\'e}, Poth, Kamath,
  Vuli{\'c}, Ruder, Cho, and Gurevych}]{pfeiffer-etal-2020-adapterhub}
Jonas Pfeiffer, Andreas R{\"u}ckl{\'e}, Clifton Poth, Aishwarya Kamath, Ivan
  Vuli{\'c}, Sebastian Ruder, Kyunghyun Cho, and Iryna Gurevych. 2020.
\newblock \href {https://doi.org/10.18653/v1/2020.emnlp-demos.7}
  {{A}dapter{H}ub: A framework for adapting transformers}.
\newblock In \emph{Proceedings of the 2020 Conference on Empirical Methods in
  Natural Language Processing: System Demonstrations}, pages 46--54, Online.
  Association for Computational Linguistics.

\bibitem[{Poth et~al.(2023)Poth, Sterz, Paul, Purkayastha, Engländer, Imhof,
  Vulić, Ruder, Gurevych, and Pfeiffer}]{poth2023adapters}
Clifton Poth, Hannah Sterz, Indraneil Paul, Sukannya Purkayastha, Leon
  Engländer, Timo Imhof, Ivan Vulić, Sebastian Ruder, Iryna Gurevych, and
  Jonas Pfeiffer. 2023.
\newblock \href {http://arxiv.org/abs/2311.11077} {Adapters: A unified library
  for parameter-efficient and modular transfer learning}.

\bibitem[{Rebuffi et~al.(2017)Rebuffi, Bilen, and Vedaldi}]{NIPS2017_e7b24b11}
Sylvestre-Alvise Rebuffi, Hakan Bilen, and Andrea Vedaldi. 2017.
\newblock \href
  {https://proceedings.neurips.cc/paper_files/paper/2017/file/e7b24b112a44fdd9ee93bdf998c6ca0e-Paper.pdf}
  {Learning multiple visual domains with residual adapters}.
\newblock In \emph{Advances in Neural Information Processing Systems},
  volume~30. Curran Associates, Inc.

\bibitem[{Scudder(2022)}]{scudder2022measuring}
Mary~F Scudder. 2022.
\newblock \href {https://doi.org/https://doi.org/10.1177/1065912921989449}
  {Measuring democratic listening: A listening quality index}.
\newblock \emph{Political research quarterly}, 75(1):175--187.

\bibitem[{Shin and Rask(2021)}]{su13031187}
Bokyong Shin and Mikko Rask. 2021.
\newblock \href {https://doi.org/10.3390/su13031187} {Assessment of online
  deliberative quality: New indicators using network analysis and time-series
  analysis}.
\newblock \emph{Sustainability}, 13(3).

\bibitem[{Steenbergen et~al.(2003)Steenbergen, B{\"a}chtiger, Sp{\"o}rndli, and
  Steiner}]{steenbergen2003measuring}
Marco~R Steenbergen, Andr{\'e} B{\"a}chtiger, Markus Sp{\"o}rndli, and J{\"u}rg
  Steiner. 2003.
\newblock \href {https://doi.org/https://doi.org/10.1057/palgrave.cep.6110002}
  {Measuring political deliberation: A discourse quality index}.
\newblock \emph{Comparative European Politics}, 1:21--48.

\bibitem[{Tausczik and Pennebaker(2010)}]{doi:10.1177/0261927X09351676}
Yla~R. Tausczik and James~W. Pennebaker. 2010.
\newblock \href {https://doi.org/10.1177/0261927X09351676} {The psychological
  meaning of words: Liwc and computerized text analysis methods}.
\newblock \emph{Journal of Language and Social Psychology}, 29(1):24--54.

\bibitem[{Vaswani et~al.(2017)Vaswani, Shazeer, Parmar, Uszkoreit, Jones,
  Gomez, Kaiser, and Polosukhin}]{NIPS2017_3f5ee243}
Ashish Vaswani, Noam Shazeer, Niki Parmar, Jakob Uszkoreit, Llion Jones,
  Aidan~N Gomez, \L~ukasz Kaiser, and Illia Polosukhin. 2017.
\newblock \href
  {https://proceedings.neurips.cc/paper_files/paper/2017/file/3f5ee243547dee91fbd053c1c4a845aa-Paper.pdf}
  {Attention is all you need}.
\newblock In \emph{Advances in Neural Information Processing Systems},
  volume~30. Curran Associates, Inc.

\bibitem[{Wilms et~al.(2023)Wilms, Stoll, Ziegele, and Gerl}]{wilms2023}
Lena Wilms, Anke Stoll, Marc Ziegele, and Katharina. Gerl. 2023.
\newblock {Bildungsbezogene Biases in crowd-annotierten Daten zur automatischen
  Klassifikation von konstruktiven und inzivilen Kommentaren (Educational
  biases in crowd-annotated data for the automatic classification of
  constructive and incivil comments)}.
\newblock In \emph{Annual Conference of the Political Communication Devision of
  the German Association of Communication Science (DGPuK)}.

\bibitem[{Wyss et~al.(2015)Wyss, Beste, and B{\"a}chtiger}]{wyss2015decline}
Dominik Wyss, Simon Beste, and Andr{\'e} B{\"a}chtiger. 2015.
\newblock \href {https://doi.org/https://doi.org/10.1111/spsr.12179} {A decline
  in the quality of debate? the evolution of cognitive complexity in swiss
  parliamentary debates on immigration (1968--2014)}.
\newblock \emph{Swiss Political Science Review}, 21(4):636--653.

\bibitem[{Ziegele et~al.(2020)Ziegele, Quiring, Esau, and
  Friess}]{doi:10.1177/0093650218797884}
Marc Ziegele, Oliver Quiring, Katharina Esau, and Dennis Friess. 2020.
\newblock \href {https://doi.org/10.1177/0093650218797884} {Linking news value
  theory with online deliberation: How news factors and illustration factors in
  news articles affect the deliberative quality of user discussions in sns’
  comment sections}.
\newblock \emph{Communication Research}, 47(6):860--890.

\end{thebibliography}


\end{document}